\documentclass[final]{cvpr}

\usepackage{times}
\usepackage{epsfig}
\usepackage{graphicx}
\usepackage{amsmath}
\usepackage{amssymb}
\usepackage{tabularx}
\usepackage{multirow}
\usepackage{subfig}
\usepackage{color,soul}
\usepackage[numbers,sort]{natbib}

\usepackage[pagebackref=true,breaklinks=true,colorlinks,bookmarks=false]{hyperref}

\def\cvprPaperID{17} 
\def\confYear{CVPR 2021}

\begin{document}

\title{Puck localization and multi-task event recognition in broadcast hockey videos}

\author{Kanav Vats  \quad  \text{Mehrnaz Fani}  \quad \text{David A. Clausi} \quad  \text{John Zelek}\\
University of Waterloo \hspace{2cm}\\
Waterloo, Ontario, Canada\\
{\tt\small \{k2vats, mfani,  dclausi,jzelek\}@uwaterloo.ca}}

\maketitle

\begin{abstract}
Puck localization is an important problem in ice hockey video analytics useful for analyzing the game, determining play location, and assessing puck possession. The problem is challenging due to the small size of the puck, excessive motion blur due to high puck velocity and occlusions due to players and boards.  In this paper, we introduce and implement a network for  puck localization in broadcast hockey video. The network leverages expert NHL play-by-play annotations and uses temporal context to locate the puck. Player locations are incorporated into the network through an attention mechanism by encoding player positions with a Gaussian-based spatial heatmap drawn at player positions.  Since event occurrence on the rink and puck location are related, we also perform event recognition by augmenting the puck localization network  with an event recognition head and training the network through multi-task learning. Experimental results demonstrate that the network is able to localize the puck with an AUC of $73.1 \%$ on the test set. The puck location can be inferred in 720p broadcast videos at $5$ frames per second. It is also demonstrated that multi-task learning with puck location improves event recognition accuracy. 
\end{abstract}

\section{Introduction}

Ball tracking in sports is of immense importance to coaches, analysts, athletes and fans. The location of the ball is directly related with the location of the play and can also be used in tasks such as player and team possession analysis. Hence, a computer vision based ball tracking/localization system can be of high utility. Although there has been significant effort for soccer ball tracking \cite{yamada, ariki,wang, ishii}, hockey puck tracking is more challenging due to a puck's small size, velocity, and regular occlusion due to players and opaque boards. \\

\begin{figure}[t]
\begin{center}
\includegraphics[width=\linewidth, height=3cm]{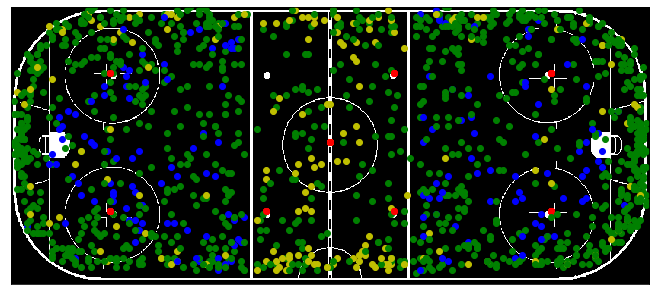}
\end{center}
  \caption{Subset of 1500 puck locations in the dataset. The puck locations on the ice rink are highly correlated with the event label. Faceoffs(red) are located at the faceoff circles, shots(blue) are located in the offensive zones and dump in/outs (yellow) are presents in the neutral zone.}
\label{fig:puck_loc}
\end{figure}

Many authors either only track the ball in screen coordinates \cite{zhang_golf,reno, Komorowski2019DeepBallDN} or track ball on the  field by treating it as a two-stage process: (1) tracking the ball in the screen coordinates  (2) registering the screen coordinates to the field coordinates using  automated homography  \cite{Jiang_2020_WACV, sharma} after performing tracking.  A big issue in ball tracking is the requirement of a large amount of frame-by-frame ball annotations for training which can be very difficult and time consuming to obtain \cite{pida}. \\

\begin{figure*}[t]
\begin{center}
\includegraphics[width = \linewidth, height = 5.2cm]{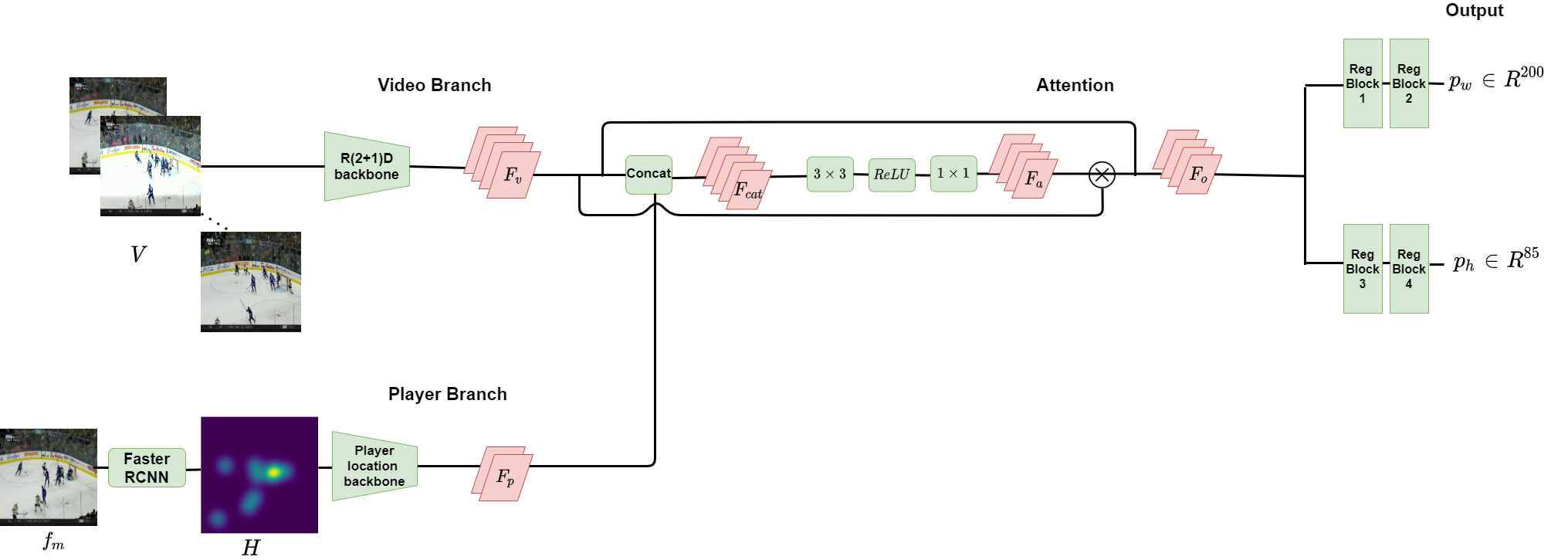}
\end{center}
  \caption{The overall network architecture. Green represents model layers while pink represents intermediate features. The network consists of four components: (1) Video Branch, (2) Player Branch, (3) Attention, and (4) Output.  The Video Branch extracts spatio-temporal features from raw hockey video. The Player Branch extracts play location information from player Gaussian heatmaps. The Attention component fuses the player location and spatio-temporal video information. The Output component produces the puck location output from the features obtained from the attention component. }
\label{fig:overall_arch}
\end{figure*}

In this paper, we introduce a  successful network for localizing hockey puck on the ice rink. The model directly estimates the puck location on the ice rink (instead of the afore-mentioned two-stage approach). Rather than estimating puck location from static images, the model estimates the puck location from video using the temporal context and leverages player location information with heatmaps
using an attention mechanism (Fig. \ref{fig:overall_arch}). Instead of annotating data on a frame-by-frame basis, we utilize the existing NHL data available on a play-by-play basis annotated by expert annotators. Experimental results demonstrate that the network is able to locate the puck with an AUC of $73.1\%$ on the test set. The network is able to localize the puck during player and board occlusions.  At test-time, the network is able to perform inference using a sliding window approach in previously unseen untrimmed broadcast hockey video at  $5$ frame per second (fps).\\


Player and puck location information is related with event occurring on the rink (Fig. \ref{fig:puck_loc}). Other research leverages player and ball trajectories for event recognition using a separate tracking/localization system \cite{Mehrasa2017LearningPT, Sanford_2020_CVPR_Workshops}.  We attach an event recognition head to the puck localization model to leverage the puck location information for event recognition and train the whole network using multi-task learning. Experimental results demonstrate that event recognition accuracy can be improved using puck location information as an additional signal.  \\

\section{Background}


\subsection{Ball tracking using traditional computer vision}

In soccer, a common approach to the ball tracking problem is a two-stage approach  \cite{yamada, ishii}: (1) ball tracking in screen coordinates and (2) sports field registration via homography.  Yamada \textit{et al.} \cite{yamada} perform camera calibration by matching straight and curved lines in the soccer field coordinates to the model. Candidates for the ball are identified by looking for white patches and tracking is performed with the help of a 3D motion model.  Ishii \textit{et al.} \cite{ishii} use a two synchronized camera system to track the soccer ball in 3D coordinates with ball detection done through template matching and tracking is done with the help of a 3D Kalman filter. Ariki \textit{et al.} \cite{ariki} use a combination of global and local search for soccer ball tracking, with the global search consisting of template matching and local approach consisting of a particle filter. Yu \textit{et al.} \cite{yu} propose a trajectory based algorithm for ball tracking in tennis where instead of determining whether an object candidate is the ball, trajectory candidates are classified into ball trajectories. Wang \textit{et al.} \cite{wang} propose a unique conditional random field (CRF) based algorithm to exploit the contextual relationship between the players and ball for ball tracking. Yakut \textit{et al.} \cite{yakut} used background subtraction to track hockey puck in zoomed in broadcast videos for short time intervals. The puck tracking performance deteriorated with high motion blur, fast camera motion and occlusions.  \\

\subsection{Ball tracking using deep learning}

Recently, deep neural networks (DNNs) have found application in sports ball tracking. Zhang \textit{et al.} \cite{zhang_golf} track golf ball in high resolution, slow-motion videos using a patch based object detector and discrete Kalman filter. Komorowski \textit{et al.} \cite{Komorowski2019DeepBallDN} use a fully convolutional network utilizing multiscale features to predict soccer ball confidence maps. Reno \textit{et al.} \cite{reno} use a convolutional neural network (CNN) with image patches as input to detect the presence of tennis balls. Our work is related to Voeikov \textit{et al.} \cite{Voeikov2020TTNetRT} where they introduce a multi-task approach for tracking a table-tennis ball using a cascade of detectors using frame-level ball location annotations. \par

Puck tracking in hockey is relatively unexplored due to the high level of difficulty involved. Pidaparthy \textit{et al.} \cite{pida} propose using a CNN to regress the puck's pixel coordinates from single high-resolution frames collected via a static camera for the purpose of automated hockey videography. The method involved an extensive annotation pipeline for model training. Instead of inferring the ball location from images and frame level annotations, we use a CNN to predict the puck location on the ice rink directly from short videos with approximate annotations without using any external homography model.

\subsection{Event recognition in sports}

In the literature, video understanding in sports is often framed as event spotting, aimed at associating events with anchored time stamps \cite{Giancola_2018_CVPR_Workshops,McNally_2019_CVPR_Workshops}, player level action recognition \cite{vats_ar, fani}  and event recognition which  involves directly classifying a video into one of the known categories. \cite{Tora_2017_CVPR_Workshops, Piergiovanni_2019_CVPR_Workshops}.  Event recognition is an important task in vision-based sports video analytics. Tora \textit{et al.} \cite{Tora_2017_CVPR_Workshops} recognize hockey event from video by gathering player level contextual interaction with the help of an LSTM. Others make use of pre-computed player and ball trajectories for recognizing events \cite{Mehrasa2017LearningPT, Sanford_2020_CVPR_Workshops}.  Mehrsa \textit{et al.} \cite{Mehrasa2017LearningPT} use player trajectories obtained from a player tracking system in order to utilize them for event recognition as well as team-classification in ice hockey. Sanford \textit{et al.} \cite{Sanford_2020_CVPR_Workshops} use player and ball trajectories obtained from a tracking system for detecting events in soccer. Instead of using player trajectories, we use puck location information to recognize hockey events using multi-task learning.

\section{Methodology}
\label{section:methodology}
\subsection{Dataset}
\begin{figure}[t]
\begin{center}
\includegraphics[width=4cm, height=4cm]{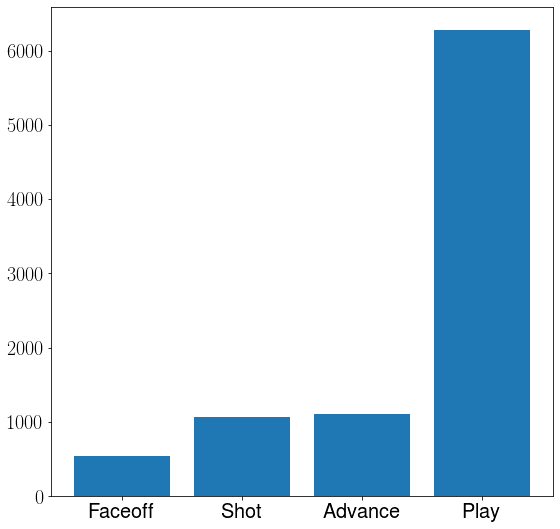}
\end{center}
  \caption{Distribution of event labels in the dataset. The dataset is imbalanced with Play event having the most occurrence.}
\label{fig:event_dist}
\end{figure}

The dataset consists  8,987 broadcast NHL videos of two second duration with a resolution of 1280 $\times$
720 pixels and a framerate of 30 fps with the approximate puck location on the ice rink annotated. The annotations are rough and approximate such that the puck location corresponds to the whole two second video clip rather than a particular frame. The videos are split into 80\% samples for training and 10\% samples each for validation and testing. Fig ~\ref{fig:puck_loc} shows the distribution of a subset of puck location data. The videos are also annotated with an event label which can be either Faceoff, Advance (dump in/out), Play ( player moving the puck with an intended recipient e.g., pass, stickhandle ) or Shot. The distribution of event labels is shown in Fig. ~\ref{fig:event_dist}.


\begin{table}[!t]
    \centering 
    \caption{Network architecture of player location backbone.  k,s and p denote kernel dimension, stride and padding respectively. $Ch_{i}$, $Ch_{o}$ and $b$ denote the number of channels going into and out of a block and batch size respectively.Additionally each layer contained a residual-skip connection with a $1\times1$ convolution. }
    \footnotesize
    \setlength{\tabcolsep}{0.2cm}
    \begin{tabular}{c}
    \hline \textbf{Input: Player heatmap} $b\times256\times256$  \\\hline\hline
     \textbf{Layer 1}\\
     Conv2D \\
 $Ch_{i} = 1, Ch_{o} = 2$ \\
   (k = $3\times3$,
      s = $2$,
   p = 1) \\
      Batch Norm 2D    \\
     ReLU  \\ \hline
          \textbf{Layer 2}\\
     Conv2D \\
 $Ch_{i} = 2, Ch_{o} = 4$ \\
   (k = $2\times2$,
      s = $2$,
   p = 0) \\
      Batch Norm 2D    \\
     ReLU  \\
 \hline
 \textbf{Layer 3} \\
 Conv2D \\
 $Ch_{i} = 4, Ch_{o} = 8$ \\
   (k = $2 \times 2$,
      s = $2$,
   p = 0) \\
      Batch Norm 2D    \\
     ReLU 
    \\ \hline
        \textbf{Output} $b\times32\times32\times8$ \\ \hline
    \end{tabular}
    \label{table:player_branch_network}
\end{table}

\begin{table}[!t]
    \centering 
    \caption{Network architecture of Regblocks 1  and 2 for output $p_w \in R^{200}$.  k,s and p denote kernel dimension, stride and padding respectively. $Ch_{i}$, $Ch_{o}$ and $b$ denote the number of channels going into and out of a block and batch size respectively. Additionally each layer contained a residual-skip connection with a $1\times1\times1$ convolution.}
    \footnotesize
    \setlength{\tabcolsep}{0.2cm}
    \begin{tabular}{c}
    \hline \textbf{Input: $F_0$} $b\times4 \times32\times32\times256$  \\\hline\hline
     \textbf{Reg Block 1}\\
     Conv3D \\
 $Ch_{i} = 256, Ch_{o} = 200$ \\
   (k = $2\times2\times2$,
      s = $2\times2\times2$,
   p = 0) \\
      Batch Norm 3D    \\
     ReLU  \\ \hline
          \textbf{Reg Block 2}\\
     Conv3D \\
 $Ch_{i} = 200, Ch_{o} = 200$ \\
   (k = $2\times2\times2$,
      s = $2\times2\times2$,
   p = 0) \\
      Batch Norm 3D    \\
     ReLU  \\
 \hline
Global average pooling
    \\ \hline
    Sigmoid activation \\
   \hline
        \textbf{Output} $b\times200$ \\ \hline
    \end{tabular}
    \label{table:output_pw}
\end{table}

\subsection{Puck localization}
The overall network architecture consists of four components: Video branch, Player branch, Attention and Output. The architecture is illustrated in Fig. \ref{fig:overall_arch}. The next four subsections explain the components in detail. 
\\
\subsubsection{Video branch}
The purpose of the video branch is to obtain relevant spatio-temporal information to estimate puck location. The video branch takes as input $16$ frames $\{f_i \in R^{256\times 256 \times 3}, i \in \{1..16\}\}$ sampled from a short video clip $V$ of two second duration.  The frames are passed through a backbone network consisting of four layers of R(2+1)D network \cite{r21d} to obtain features $F_v \in R^{4\times 32 \times 32 \times 256}$ to be used for further processing. The R(2+1)D network consists of (2+1)D blocks which splits spatio-temporal convolutions into spatial 2D convolutions followed by a temporal 1D convolution.  \\

\subsubsection{Player branch}

The location of puck on the ice rink is correlated with the location of the players since the puck is expected to be be present where the player "density" is high. We make the assumption that the location of players remains approximately the same in a short two second video clip. In order to encode the spatial player location, we take the middle frame $f_m$ of the video $V$ and pass it through a FasterRCNN \cite{fasterrcnn} network to detect  players. After player detection, we draw a Gaussian with a standard deviation of $\sigma_p$ at the centre of the player bounding boxes to obtain the player location heatmap $H$. An advantage of using this representation is that the player location variability in the video clip can be expressed through the Gaussian variance. The player location heatmap $H$ is passed through a player location backbone network to output player location features $F_p \in R^{32 \times 32 \times 8}$ . The exact configuration of the player location backbone is shown in Table \ref{table:player_branch_network}. The player location features $F_p$ are passed to the attention block for further processing. \\

\subsubsection{Attention}
The purpose of attention is to make the network incorporate player locations by considering the relationship between video features $F_v$ and player location features $F_p$. The player location features $F_p$ and video features $F_v$ are concatenated along the the channel axis by repeating the player location features along the temporal axis. The concatenated features  $F_{cat} \in R^{4\times 32 \times 32 \times 264}$ are then passed through a variation of the squeeze and excitation \cite{squeeze, bhuiyan} network consisting of a $3\times3$ convolution, non-linear excitation and $1\times1$ convolution. The $3\times3$ squeeze operation learns the spatial relationships between  player locations on the rink and video features. The squeeze operation outputs features  $F_{cat}^\prime \in R^{4\times 32 \times 32 \times 132}$. The squeeze operation is followed by non linear activation and $1\times1$ convolution to obtain features $F_a \in  R^{4\times 32 \times 32 \times 256} $. The $1\times1$ convolution learns the channel wise relationships between the feature maps in $F_{cat}^\prime$. Finally, the output of the attention block is the hadamard product of the attention features $F_a$ and the video features $F_{v}$ followed by a skip connection.
     \begin{equation} F_o= F_a \otimes F_{v} + F_{v}
     \end{equation}
\subsubsection{Output}
The features $F_o$ obtained from the attention component are finally passed through two RegBlocks to output the probability of puck location on the ice rink. Global average pooling is done at the end of the two RegBlocks to squash the intermediate output to one dimensional vectors.  This is done independently for rink width and height dimensions through two separate branches. The overall network outputs two vectors, $p_w \in R^{200}$ and $p_h \in R^{85}$,  in accordance with the dimension of the NHL rink. The exact details of RegBlocks 1 and 2 are shown in Table \ref{table:output_pw}. Regblocks 3 and 4 have a similar architecture, the only difference is that instead of a $R^{200}$ vector $p_{w}$, a $R^{85}$ vector $p_{h}$ is output by changing the output channels to 85.  \\

\subsubsection{Training details}
\label{training_details}
\begin{figure}[t]
\begin{center}
\includegraphics[width=\linewidth, height = 4cm]{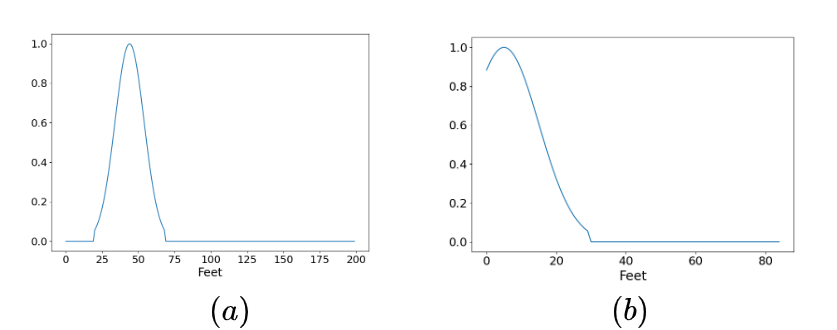}
\end{center}
  \caption{Construction of ground truth for a training sample with puck located at $w=44 \; ft$ and $h=5 \; ft$. (a) Ground truth distribution vector $w_{gt} \in R^{200}$ (b) Ground truth distribution vector $h_{gt} \in R^{85}$}
\label{fig:make_gt}
\end{figure}

We use the cross entropy loss to train the network. In order to create the ground truth, we use a one dimensional Gaussian with mean at the ground truth puck location and a standard deviation $\sigma$ for both directions. The Gaussian variance encodes the variability in ball location in the short video clip (Fig. \ref{fig:make_gt}) .  The total loss  $L_{puck}$ is the sum of the loss in horizontal axis $L_{w}$ and vertical axis  $L_{h}$, which is given by:
\begin{equation}
     L_{puck} = L_{w} + L_{h} 
     \label{eq2}
\end{equation}
\begin{equation}
    L_{puck} = -\frac{1}{200}\sum_{i=1}^{200}w_{gt}\log{p_w} -\frac{1}{85}\sum_{j=1}^{85}h_{gt}\log{p_h}
\end{equation}
Where $w_{gt} \in R^{200}$ and $h_{gt} \in R^{85}$ denote the ground truth probabilities and $p_w \in R_{200}$ and $p_h \in R^{85}$ denote the predicted probabilities.\\
For data augmentation, each frame is sampled from a uniform distribution $U(0,60)$ so that the network sees different frames of the same video when the video sampled different times.  The data augmentation technique is used is all experiments unless stated otherwise. We use the Adam optimizer with an initial learning rate of .0001 such that the learning rate is reduced by a factor of $\frac{1}{5}$ at iteration number 5000. The batch size is 15.

\begin{figure}[t]
\begin{center}
\includegraphics[width=\linewidth]{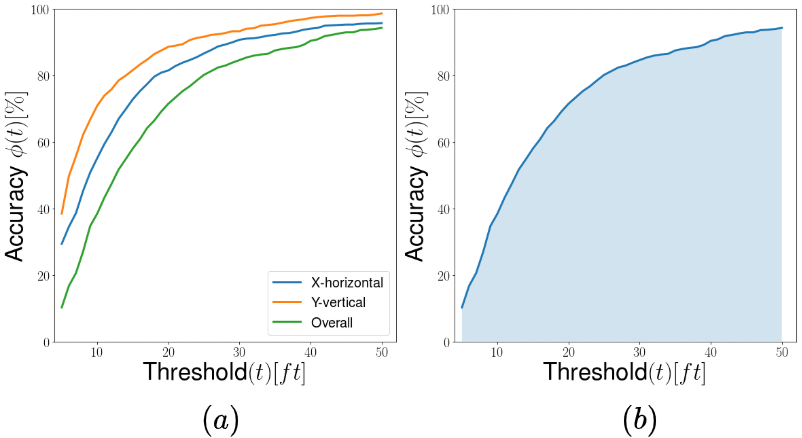}
\end{center}
  \caption{ (a) Accuracy ($\phi$) vs threshold ($t$) curve. (b) The best performing model gets an overall AUC of $73.1\%$ on test set.}
\label{fig:auc_area}
\end{figure}


\subsection{Multi-task event recognition}
The event occurring on the rink in hockey is highly correlated with the puck location. For example, faceoff occurs on the faceoff circles, shots are expected to occur in the offensive zones etc. In order to leverage the shared information between puck location and event recognition, we learn the event and puck location in hockey video clip using a single network through multi-task learning. This is done by appending a third event recognition head at the end of features $F_o$ representing the probability of the predicted event $p_e \in R^4$. Let $Ch_{i}$, $Ch_{o}$ and $k$ denote the number of channels going into and out of a kernel and kernel size respectively. The event recognition head consists of a 3D convolution layer with $Ch_{i} =256$, $Ch_{o} =256$ with $k = 2\times3\times3$ and $stride =2$ followed by 3D batch normalization , followed by another 3D convolution  $Ch_{i} =256$, $Ch_{o} = 512$ with $k = 2\times3\times3$ and $stride =2$, adaptive pooling and fully connected layer. The total loss is the linear combination of equation \ref{eq2} and the event loss $L_e$ which is a cross entropy loss between the ground truth and predicted event probability.
Following Cipolla \textit{et al.} \cite{mtl}, the overall loss for the muti-task network is given by:

\begin{equation}
    L_{multi} =  \frac{1}{\sigma_1^2}L_{w} +  \frac{1}{\sigma_2^2} L_{h} +  \frac{1}{\sigma_3^2} L_{e} + \log(\sigma_1) + \log(\sigma_2) + \log(\sigma_3)
\end{equation}

The rest of the training details and data augmentation are the same as in section \ref{training_details}.

\section{Results}
\label{section:results}
\subsection{Puck localization}
\subsubsection{Accuracy metric}
A test video is considered to be correctly predicted at a tolerance $t$ feet if the distance between the ground truth puck location $z$ and predicted puck location $z_{p}$ is less than $t$ feet. That is $||z - z_{p}||_{2}<t$. Let $\phi(t)$ denote the percentage of examples in the test set with correctly predicted position puck position at a tolerance of $t$. We define the accuracy metric as the area under the curve (AUC) $\phi(t)$   at tolerance of $t=5$ feet to $t=50$ feet.

\subsubsection{Trimmed video clips}

The network attains an AUC of $73.1\%$ on the test dataset illustrated in Fig. \ref{fig:auc_area} (b). The AUC in the horizontal direction is $81.4\%$ and AUC in vertical direction is $87.8\%$. From Fig. \ref{fig:auc_area} (a), at a low tolerance of $t=12 \, ft$, the accuracy in vertical(Y) direction is $76\%$ and the accuracy in horizontal(X) direction is $63\%$. At a tolerance of $t=20 \, ft$, the accuracy in both directions is greater than $80\%$ . 

\begin{figure}[t]
\begin{center}
\includegraphics[width=\linewidth,height = 3cm]{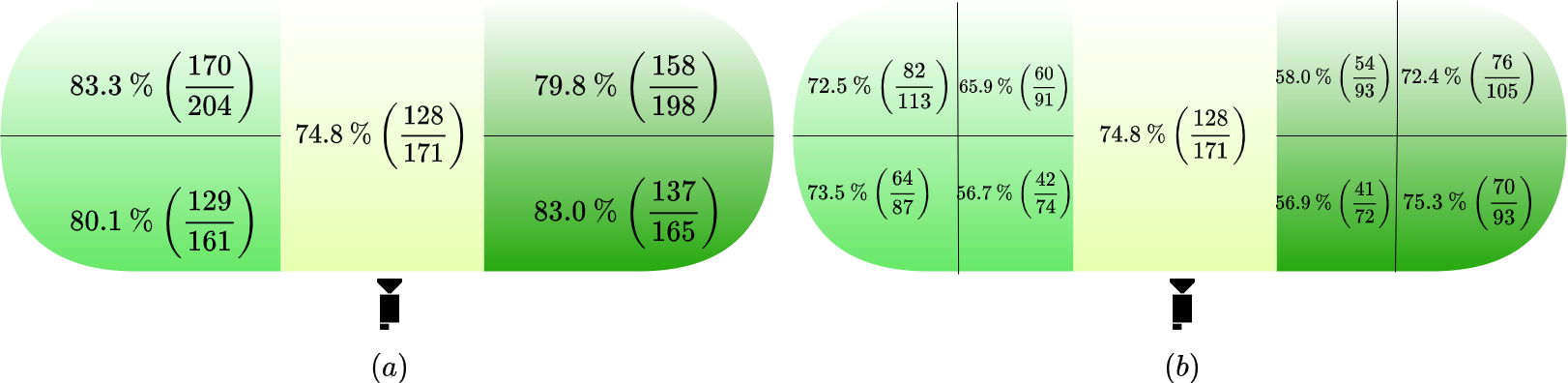}
\end{center}
  \caption{Zone-wise accuracy. The figure represents the
hockey rink with the text in each zone represents the percentage of test examples predicted correctly in that zone.
The position of the camera is at the bottom. In (b), the accuracy is low in the lower halves of the defensive and offensive zones since the puck gets occluded by the rink board.}
\label{fig:zone_2}
\end{figure}

Fig.  \ref{fig:zone_2} show the zone wise accuracy. A test example is classified correctly if the predicted and ground truth puck location lies in the same zone. From Fig. \ref{fig:zone_2} (a), the network gets an  accuracy of $\sim 80\%$ percent in the upper and lower halves of the offensive and defensive zones. From Fig. \ref{fig:zone_2} (b), after further splitting the ice rink in nine zones, the network achieves an accuracy of more than $70\%$ in five zones. The network also has failure cases. From Fig. \ref{fig:zone_2} (b), it can be seen that accuracy is low (less than $60 \%$ ) in the bottom halves of the defensive and offensive zones. This is due to the puck being occluded by the rink boards. \\

\subsubsection{Untrimmed broadcast video}

We also test the network on untrimmed broadcast videos using a sliding window of length $l$ and stride $s$. The window length $l$ is the time duration covered by the sliding window and stride $s$ is the time difference between two consecutive application of the sliding window. Due to the difficulty of annotating puck location frame-by-frame in $720p$ videos, we do not possess the frame-by-frame ground truth puck location.  Therefore, we perform a qualitative analysis in this section. The videos used for testing are previously unseen video not present in the dataset used for training and testing the network.  \\

\begin{figure*}[t]
\begin{center}
\includegraphics[width=\linewidth, height = 4.5cm]{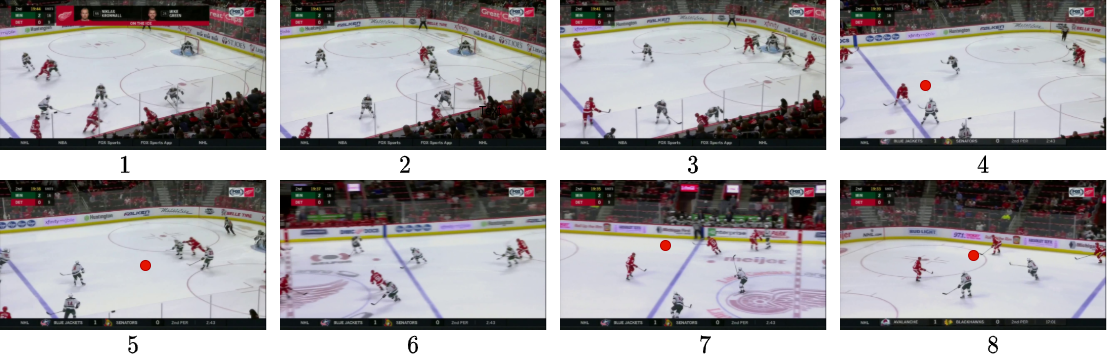}
\end{center}
  \caption{ Some frames from the 10 second validation video clip. Whenever visible, the location of the puck is highlighted using the red circle. The initial portion of the clip is challenging since the puck is not visible in the initial part of the clip.}
\label{fig:first_vid}
\end{figure*}

\begin{figure}[t]
\begin{center}
\includegraphics[width=\linewidth, height = 3cm]{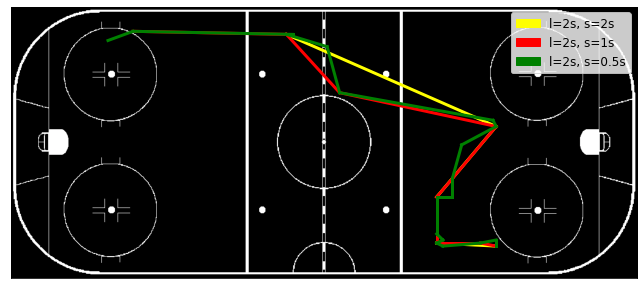}
\end{center}
  \caption{Puck trajectory on the ice rink for the validation video. The trajectory becomes noisy with $s=0.5s$ and lower.}
\label{fig:traj_val}
\end{figure}

\begin{figure}[t]
\begin{center}
\includegraphics[width=\linewidth, height = 3cm]{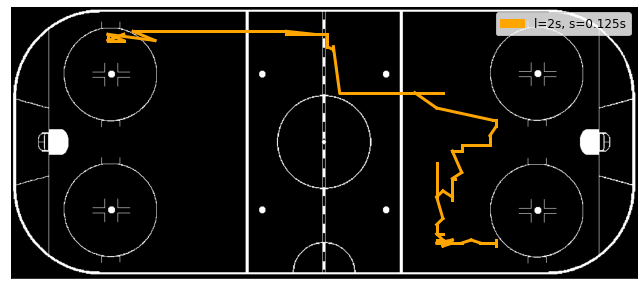}
\end{center}
  \caption{Puck trajectory for the validation video with a very low stride of $0.125$ seconds. The trajectory is extremely noisy and hence is not a good estimate.}
\label{fig:low_stride}
\end{figure}

\begin{figure*}[t]
\begin{center}
\includegraphics[width=\linewidth, height = 4cm]{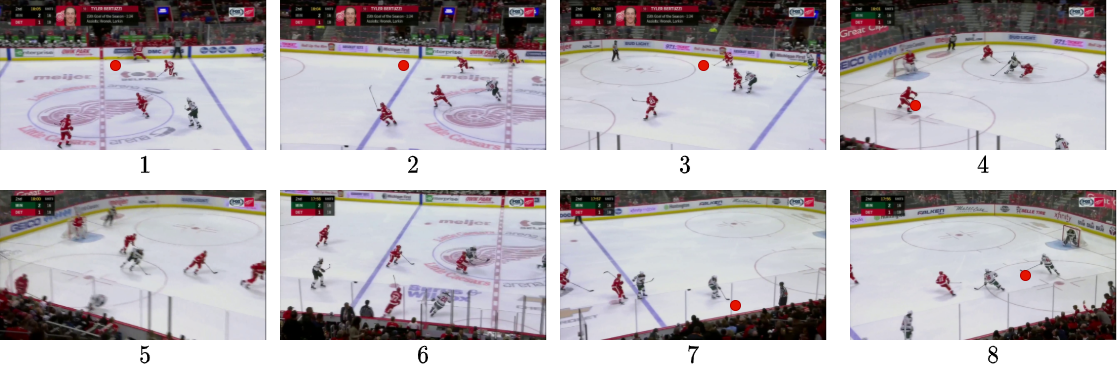}
\end{center}
  \caption{ Some frames from the test 10 second clip. Whenever visible, the location of the puck is highlighted using the red circle}
\label{fig:second_vid}
\end{figure*}

To determine the optimal values of stride $s$ validation is performed on a 10 second clip.  Some frames from the validation 10 second clip are shown in  Fig. \ref{fig:first_vid}. Whenever visible, the location of the puck is highlighted using a red circle. Fig \ref{fig:traj_val} shows the trajectories obtained. The network is able to approximately localize the puck in untrimmed video within acceptable visual errors, even though the network is trained on trimmed video clips where puck location is annotated approximately. The puck is not visible during many frames of the video, but the network is still able to guess the puck location. This is because the network takes into account the temporal context and player location. Since the network is originally trained on $2$ second clips, the window length $l$ is fixed to $2s$. Fig \ref{fig:traj_val} , shows that as the stride $s$ is decreased, the puck location estimates become noisy. Since between two passes, the puck motion is  linear, we do not decrease stride below $0.5s$ as it leads to very noisy estimates (Fig. \ref{fig:low_stride}). The optimal stride $s=1s$  gives the most accurate result. A lower stride results in noisy results and higher strides produces very simple predictions. \\
The network is  tested on another 10 second video  with $l=2s$ and $s=1s$ shown in Fig \ref{fig:second_vid}. The predicted puck trajectory is shown in Fig \ref{fig:second_vid}. The puck is occluded by the rink board during a part of the video (shown in images 5 and 6). The network is able to localize the puck even when it is not visible due to board occlusions.The inference time of the network on a single GTX 1080Ti GPU with 12GB memory is $5$ fps.

\subsection{Ablation studies}
We perform an ablation study on the  number of layers in the backbone network, puck ground truth standard deviation, presence/absence of player branch  consisting of player locations and data augmentation .

\subsubsection{Puck ground truth standard deviation}
The best value of standard deviation $\sigma$ of puck location ground truth 1D Gaussian is determined by varying $\sigma$ from 20 to 35 in multiples of five. From Table \ref{table:sigma_gt}, the number of layers in the backbone is fixed to three while player location based attention is not used. Maximum AUC of $69\%$ is attained with $\sigma=30$ feet. A lower value of $\sigma$ makes the ground truth Gaussian more rigid/peaked which makes learning difficult. A value of sigma greater than $30$ lowers accuracy since a higher $\sigma$ makes the ground truth more spread out which reduces accuracy on lower tolerance values.

\subsubsection{Layers in backbone}
We determine the optimal number of layers in the R(2+1)D backbone network by extracting the video branch features from different layers without using the player location based attention.  The puck ground truth standard deviation is set to the optimal value of $30$. From Table \ref{table:backbone_layers}, the maximum AUC of $72.5\%$ is achieved by using 4 layers of R(2+1)D network. Further increasing the number of backbone layers to 5 causes a decrease of $0.1$ in AUC due to overfitting.\\

\begin{table}[!t]

    \centering
    \caption{Comparison of AUC with different values of $\sigma$ with a three layer backbone network. Network with $\sigma = 30$ shows the best performance }
    \footnotesize
    \setlength{\tabcolsep}{0.2cm}
    \begin{tabular}{c|c|c|c}\hline
  
       $\sigma$ & AUC & AUC(X) & AUC(Y)\\\hline
       $20$  &  $62.5$ & $71.3$ & $85.07$\\
      $25$ &  $68.5$ & $77.9$ & $85.6$\\
    $30$ & $\textbf{69.0}$ & $78.5$ & $85.5$\\
       $35$ &  $68.9$ & $78.8$ & $85.4$ \\ 
    \end{tabular}
    \label{table:sigma_gt}
\end{table}

\begin{table}[!t]

    \centering
    \caption{Comparison of AUC with different number of layers of the backbone R(2+1)D network. A four layer backbone shows the best performance. }
    \footnotesize
    \setlength{\tabcolsep}{0.2cm}
    \begin{tabular}{c|c|c|c}\hline
  
       Layers & AUC & AUC(X) & AUC(Y)\\\hline
          $2$  &   $56.3$ & $73.2$ & $74.1$\\
       $3$  & $69.0$ & $78.5$ & $85.5$\\
      $4$ &  $\textbf{72.5}$ & $81.3$ & $87.3$\\
    $5$ & $72.4$ & $81.0$ & $87.3$\\
    \end{tabular}
    \label{table:backbone_layers}
\end{table}

\subsubsection{Player location based attention}
We add the player branch and the attention mechanism to the network with 4 backbone layers and $\sigma =30$. Three values of  player location standard deviation $\sigma_p = \{15,20,25\}$ are tested. From Table \ref{table:player_branch}, adding the player location based attention mechanism brought an improvement in the overall AUC by $0.6\%$ with $\sigma_p = 15$. Further increasing $\sigma_p$ causes the player location heatmap to become more spread out obfuscating player location information. \\

\begin{table}[!t]

    \centering
    \caption{Comparison of AUC values with/without player branch. The player branch with $\sigma_p = 15$ shows the best performance.  }
    \footnotesize
    \setlength{\tabcolsep}{0.2cm}
    \begin{tabular}{c|c|c|c|c}\hline
  
       Player detection & $\sigma_p$ & AUC & AUC(X) & AUC(Y)\\\hline
      No & - &  $72.5$ & $81.3$ & $87.3$\\
     Yes & $15$ &  $\textbf{73.1}$ & $81.4$ & $87.8$\\
      Yes & $20$ &  $72.8$ & $81.5$ & $87.3$\\
       Yes & $25$ &  $72.2$ & $80.4$ & $87.9$\\
   
    \end{tabular}
    \label{table:player_branch}
\end{table}

\subsubsection{Data augmentation}

We compare the data augmentation technique done using randomly sampling frames from a uniform distribution (explained in Section \ref{training_details}) to sampling frames at a constant interval. From Table \ref{table:sampling}, removing random sampling decreases the overall AUC by $3.2\%$ which demonstrates the advantage of the data augmentation technique used.

\begin{figure*}
	\begin{center}
		\subfloat[]{
		    \includegraphics[width=0.5\linewidth, height= 3cm]{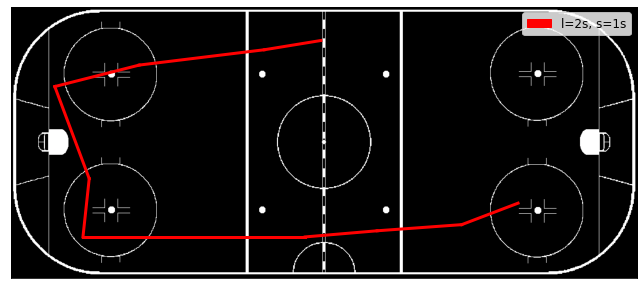}
		}
		\subfloat[]{
		    \includegraphics[width=0.5\linewidth, height= 3cm]{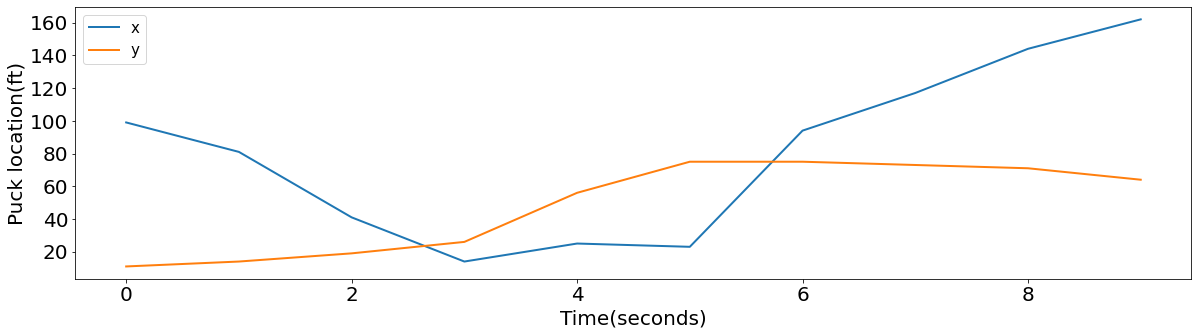}
		}
	
	\end{center}
	\caption{ The predicted puck trajectory for the test video with window length two seconds ($l = 2s$) and stride one second ($s=1s$)  . The network is able to localize the puck even when it is not visible due to board occlusions.}
	\label{fig:traj_v2}
\end{figure*}

\begin{table}[!t]

    \centering
    \caption{Comparison of AUC values with uniform and random sampling }
    \footnotesize
    \setlength{\tabcolsep}{0.2cm}
    \begin{tabular}{c|c|c|c}\hline
  
       Sampling method & AUC & AUC(X) & AUC(Y)\\\hline
        Constant interval    &   70.3  & 79.4 & 86.4\\
        Random  & $\textbf{73.1}$ & $81.4$ & $87.8$\\
          \end{tabular}
    \label{table:sampling}
\end{table}

\subsection{Multi-task event recognition}
The network performing only event recognition task with zero weights assigned to the puck location loss is treated as a comparison baseline.  We compare the macro averaged precision, recall and  F1 score values corresponding to the four events for the multi-task learning setting and the baseline. \\

From Table \ref{table:pr_values}, the multi-task setting performs better compared to the baseline where puck location is not used as an additional signal which demonstrates that learning the two tasks together is beneficial for event recognition. This is because multi-task learning with puck location provides contextual location information which greatly improves F1 score of events such as Faceoff ($82.6$ multi-task vs $78.8$ baseline) which always occur in specific rink locations. The Advance event has the lowest F1 score value of $41.9$. This is because it  often gets confused with Play and Shot events.

\begin{table}[!t]
    \centering
    \caption{Precision, Recall and F1 score values for the network corresponding to the multi-task and baseline settings. The multi-task setting shows better performance. }
    \footnotesize
    \setlength{\tabcolsep}{0.2cm}
    \begin{tabular}{c|c|c|c|c}
         & & Precision  & Recall & F1 score \\\hline
        \multirow{5}{*}{Muti task} &Play & \textbf{81.8} & 87.2 & 84.4\\
       & Shot & 56.4 & \textbf{60.6} & 58.4\\
       & Advance & \textbf{63.2} & \textbf{31.3} & \textbf{41.9}\\
       & Faceoff & \textbf{76.3} & \textbf{90.0} & \textbf{82.6}\\ 
         & Macro Avg. & \textbf{69.4} & \textbf{67.3} & \textbf{66.8}\\ \hline
          \multirow{5}{*}{Baseline} &Play & 81.0 & 88.6 & 84.6\\
       & Shot & 63.5 & 56.0 & 59.5\\
       & Advance & 55.4 &31.3 & 40.0\\
       & Faceoff & 75.9 & 82.0 & 78.8\\ 
         & Macro Avg. & 69.0 & 64.5 & 65.8\\ 
    \end{tabular}
    \label{table:pr_values}
\end{table}

\section{Conclusion}
\label{section:conclusion}
A model has been designed and developed to localize puck and recognize events in broadcast hockey video. The model makes use of temporal information and player locations to localize puck. We append an event recognition head to the puck localization model and train the whole network using multi-task learning. We also perform ablation studies on the model parameters and data augmentation used. We attain an AUC of $73.1 \%$ on the test set and qualitatively localize the puck in untrimmed broadcast videos. We also report an ice rink region based average accuracy of $80.2\%$ with the ice rink split into five zones and  $67.3 \%$ with the rink split into nine regions. Experimental results also demonstrates that the puck location signal aids event recognition with the multi-task learning setting improving the macro-average event recognition F1-score  by one percent.  Future work will focus on using high resolution images/videos and frame-wise puck location annotations to improve performance.

\textbf{Acknowledgments.}
This work was supported by Stathletes through the Mitacs Accelerate Program and Natural Sciences
and Engineering Research Council of Canada (NSERC). We also acknowledge Compute
Canada for hardware support

{\small
\bibliographystyle{ieee_fullname}
\bibliography{main}
}

\end{document}